\pdfoutput=1

\documentclass[11pt]{article}

\usepackage{ACL2023}

\usepackage{times}
\usepackage{latexsym}

\usepackage[T1]{fontenc}

\usepackage[utf8]{inputenc}

\usepackage{microtype}

\usepackage{inconsolata}

\usepackage{lipsum}
\usepackage{booktabs}
\usepackage{tikz}

\definecolor[named]{ACMBlue}{cmyk}{1,0.1,0,0.1}
\definecolor[named]{ACMYellow}{cmyk}{0,0.16,1,0}
\definecolor[named]{ACMOrange}{cmyk}{0,0.42,1,0.01}
\definecolor[named]{ACMRed}{cmyk}{0,0.90,0.86,0}
\definecolor[named]{ACMLightBlue}{cmyk}{0.49,0.01,0,0}
\definecolor[named]{ACMGreen}{cmyk}{0.20,0,1,0.19}
\definecolor[named]{ACMPurple}{cmyk}{0.55,1,0,0.15}
\definecolor[named]{ACMDarkBlue}{cmyk}{1,0.58,0,0.21}
\definecolor{mygreen}{RGB}{34,139,34}
\definecolor{myorange}{RGB}{255,140,0}

\newcommand{\comp}[1]{%
    \begin{tikzpicture}[baseline=0]
        \draw [fill=ACMYellow] (0,0) rectangle (1,.25);
        \draw [fill=ACMOrange] (0,0) rectangle (#1,.25);
        \draw [densely dotted] (.5,.025) -- (.5,.25);
        \pgfmathtruncatemacro{\percentage}{int(#1 * 100)}
        \node [color=white, font=\tiny] at (.25,.125) {\percentage\%};
    \end{tikzpicture}%
}
\newcommand{\flue}[1]{%
    \begin{tikzpicture}[baseline=0]
        \draw [fill=ACMLightBlue] (0,0) rectangle (1,.25);
        \draw [fill=ACMBlue] (0,0) rectangle (#1,.25);
        \draw [densely dotted] (.5,.025) -- (.5,.25);
        \pgfmathtruncatemacro{\percentage}{int(#1 * 100)}
        \node [color=white, font=\tiny] at (.25,.125) {\percentage\%};
    \end{tikzpicture}%
}
\newcommand{\overall}[1]{%
    \begin{tikzpicture}[baseline=0]
        \draw [fill=pink] (0,0) rectangle (1,.25);
        \draw [fill=magenta] (0,0) rectangle (#1,.25);
        \draw [densely dotted] (.5,.025) -- (.5,.25);
        \pgfmathtruncatemacro{\percentage}{int(#1 * 100)}
        \node [color=white, font=\tiny] at (.25,.125) {\percentage\%};
    \end{tikzpicture}%
}

\title{Model Analysis \& Evaluation for Ambiguous Question Answering}

\author{%
    Konstantinos Papakostas\thanks{\ \ Equal contributions} ~\thanks{\ \ Correspondence email: \texttt{dinos.ppk@gmail.com}} \\
    University of Amsterdam \\ \And
    Irene Papadopoulou\footnotemark[1] \\
    University of Amsterdam %
}

\begin{document}
\maketitle
\begin{abstract}
    Ambiguous questions are a challenge for Question Answering models, as they require answers that cover multiple interpretations of the original query. To this end, these models are required to generate long-form answers that often combine conflicting pieces of information. Although recent advances in the field have shown strong capabilities in generating fluent responses, certain research questions remain unanswered. Does model/data scaling improve the answers' quality? Do automated metrics align with human judgment? To what extent do these models ground their answers in evidence? In this study, we aim to thoroughly investigate these aspects, and provide valuable insights into the limitations of the current approaches. To aid in reproducibility and further extension of our work, we open-source our code \href{https://github.com/din0s/ambig_lfqa}{here}.
\end{abstract}

\section{Introduction}
\label{sec:intro}
Question Answering (QA) has been subject to great progress in the past years, largely thanks to the representational capabilities of modern architectures like the Transformer~\cite{attention}, but also due to the curation of large, high-quality datasets that enabled the effective training of these models~\cite{joshi-etal-2017-triviaqa, kwiatkowski-etal-2019-natural}.

At the same time, the presence of \textit{ambiguous questions} has been a challenging aspect of QA. In order to answer such questions, models are required to generate long answers with fluency and cohesion, which is often referred to in the literature as Long-Form Question Answering (LFQA). To tackle this, \citet{min-etal-2020-ambigqa} curated the \texttt{AmbigQA} dataset, which contains disambiguations for various questions that were present in popular benchmarks. Extending this work, \citet{asqa} selected a subset of the questions and crowd-sourced gold answers that cover all possible interpretations of each question, resulting in the \texttt{ASQA} dataset.

Recently, \citet{krishna-etal-2021-hurdles} performed a case study on \texttt{ELI5}~\cite{fan-etal-2019-eli5}, one of the largest collections available for LFQA, and pointed out several issues that complicate the development and evaluation of suitable models. In particular, the authors questioned whether the retrieved documents are considered by the generative models when producing an answer, and the correlation of the common evaluation metrics with human judgment.

In this work, we aim to investigate whether the baselines set on \texttt{ASQA} by \citet{asqa} suffer from the issues pointed out by \citet{krishna-etal-2021-hurdles}, but also to analyze the modeling choices that contribute to performance. Concretely, we set out to answer the following research questions:
\begin{enumerate} 
    \item[RQ1] Does scaling the size of the generative models affect the quality of the generated answers?
    \item[RQ2] Can an intermediate round of fine-tuning on non-ambiguous LFQA collections improve performance in ambiguous QA?
    \item[RQ3] When comparing models head-to-head, does human judgment reflect the difference in the automated evaluation metrics?
    \item[RQ4] Do models base their answers on the retrieved evidence, or could they be hallucinating?
\end{enumerate}

\section{Methodology}
\label{sec:method}
We design a standard retrieval-augmented system, to identify the dimensions that contribute to disambiguating questions and generating factual answers.

\subsection{The LFQA Pipeline}
\paragraph{\textbf{Evidence Retrieval}}
\label{sec:retrieval}
The first step of the pipeline is to identify the documents that will form the basis of the generated answers. Given a question $q$, we employ a retrieval method $R$ that fetches the top-$k$ relevant documents $\{ d_i \}_{i=1}^k$.
For a document $d_i$ to be considered relevant, it needs to cover at least one aspect of $q$. To completely resolve the ambiguity, the passages\footnote{We use the terms document and passage interchangeably.} in the index should suffice to collectively answer all aspects of a given question.

\paragraph{\textbf{Answer Generation}}
\label{sec:generation}
Once the evidence has been collected, we feed the retrieved passages to a generative model to summarize them in a concise answer that disambiguates the question at hand. As is the standard practice in contemporary literature, we opt for a sequence-to-sequence model $G$ that follows an encoder-decoder architecture, which first creates a dense representation of the concatenation of the question and the passages, and then produces the answer by attending to this latent representation:
$$
\mathrm{answer} = G([q;d_1;\ldots;d_k])
$$
with $[\cdot~;\cdot]$ being the concatenation of $\geq 2$ passages.

\subsection{Modeling Choices}
We expect more sophisticated pipelines to provide better answers for ambiguous questions, and thus we make a distinction based on the complexity of the interaction between the two components.

\paragraph{\textbf{Naive}}
We implement the \textsc{Question} baseline, which repeats the ambiguous question a few times in order to match the typical length of the answers in the dataset. This is a lower bound on the task, as we are not truly answering the question at hand.

\paragraph{\textbf{Retrieval-Only}}
In this case, we rely exclusively on a retriever to fetch the top-$k$ passages as a response to the ambiguous question.
We experiment with different values of $k$ to evaluate whether using more passages leads to an answer that covers more of the disambiguated questions.\footnote{Employing some form of result diversification has the potential to improve the performance of the retrieval component, but we leave this direction as future work.}

\paragraph{\textbf{Sequence-to-Sequence}}
A generative language model is often used to produce a concise response that summarizes all of the disambiguating answers. We analyze three scenarios:
\begin{itemize}
    \item \underline{Closed Book}: In the most extreme approach, we assume that the model is not conditioned on the results of a retriever, but rather only on the question itself, and relies on its parametric knowledge~\cite{roberts-etal-2020-much} to respond. We expect this to significantly harm performance, as the available context to provide an accurate answer is limited.
    \item \underline{Random Retrieval}: In order to verify whether the model \textit{grounds} its answers on the retrieved documents~\cite{krishna-etal-2021-hurdles}, we design a controlled scenario where we randomly sample passages from our index as evidence.
    \item \underline{Open Book}: In the most realistic setting, the model treats the top-$k$ results of a retriever as context to respond appropriately. We expect stronger retrieval methods to lead to more comprehensive answers, as the generative model will be conditioned on more diverse and relevant information.
\end{itemize}

\section{Experimental Setup}
\label{sec:experiments}
We aim to assess whether LFQA systems can generate concise answers that disambiguate the provided questions. In this section, we present the datasets and models used, as well as the evaluation metrics.

\subsection{Datasets}
\label{sec:datasets}
We use the \texttt{ASQA} dataset to train and evaluate our systems. It is a subset of the \texttt{AmbigQA} dataset, with long-form answers and additional context for each of the selected samples. More specifically, it contains 6,316 ambiguous questions, with each one being paired to a set of disambiguated questions, the corresponding short answers, and the Wikipedia passages where the answers were found. For each question, the dataset curators crowd-sourced a long-form answer that resolves the ambiguity by summarizing all short answers. The annotators provided one reference answer for all train (4,353) samples, and two for all dev (948) and test (1,015) samples.\footnote{An ambiguity can be observed even in a simple query like ``Who was the ruler in France in 1830?'', which presents a challenge due to the existence of two rulers during that period.}

Although \texttt{ASQA} is a useful resource for LFQA, when going through the dataset for preliminary analysis, we found cases where the ambiguity was to identify when ``last/this year'' refers to. We argue that training a model on such samples is counter-intuitive, as we generally assume that the information is coming from a fixed snapshot of a knowledge base. We provide a few examples that cover similar types of questions in Appendix~\ref{sec:appendix-asqa-issues}.

Additionally, given the limited size of the \texttt{ASQA} dataset, we follow one of the proposed research directions posed by \citet{asqa} and investigate the impact of intermediate fine-tuning on a larger LFQA collection, before training on \texttt{ASQA}. In particular, we use a processed version of the \texttt{ELI5} dataset\footnote{Available on HuggingFace: \href{https://huggingface.co/datasets/vblagoje/lfqa}{lfqa} \& \href{https://huggingface.co/datasets/vblagoje/lfqa_support_docs}{lfqa\_support\_docs}.} that addresses some of the issues raised by \citet{krishna-etal-2021-hurdles} (226,147 train / 3,020 dev samples), as well as the \texttt{NLGEN} set of the \texttt{MS MARCO} QA dataset (153,725 train / 12,467 dev samples).

\subsection{Models}
\label{sec:models}
The typical QA pipeline comprises a retrieval and a generative model, which can be pre-trained separately and then fine-tuned on the downstream task. Although this can be done in an end-to-end fashion, we chose to keep the retriever frozen, to avoid re-indexing of the support passages during training, and only train the generative model. We provide detailed training information in Appendix~\ref{sec:appendix-training}.

\paragraph{\textbf{Retrieval Models}}
\label{sec:retrieval-models}
We experiment with both sparse (lexical) and dense (neural) methods, in order to investigate whether the type of the question encoder has an impact on the relevancy of the retrieved passages. We chose BM25~\cite{bm25} for the former, and DPR~\cite{karpukhin-etal-2020-dense} for the latter, as they were both supported by the Pyserini~\cite{pyserini} toolkit and they constitute two of the most explored options in their corresponding fields. In both cases, we use the pre-built indices of Wikipedia provided by Pyserini to have a common knowledge base that matches the one used by the dataset curators.

\paragraph{\textbf{Generative Models}}
\label{sec:generative-models}
We use two of the most popular Transformer-based encoder-decoder models, namely T5~\cite{t5} and BART~\cite{lewis-etal-2020-bart}. More specifically, we experiment with three variants of these models, in increasing parameter count: BART-base (140M params), T5-base (220M params), and BART-large (400M params). By doing so, we aim to verify whether increasing the capacity of the generative model corresponds to an increase in answer quality.

\subsection{Metrics}
\paragraph{\textbf{Automated Evaluation}}
\label{sec:automated-metrics}
Evaluating the performance of generative models is one of the most challenging aspects for LFQA. For recall-oriented QA systems, the most common metric used is \textsc{Rouge-L}~\cite{lin-2004-rouge}, which identifies the longest common sub-sequence between the generated answer and a reference (gold) answer. Sequences are penalized proportionally to their length to prevent generating a longer output to artificially increase the overlap with the reference text. However, a recent study by \citet{krishna-etal-2021-hurdles} revealed that \textsc{Rouge-L} did not always correlate with human judgment, pointing to the need for a more diverse evaluation setup. Specifically for ambiguous LFQA, \citet{asqa} proposed two metrics to quantify the disambiguation ability of the model:
\begin{enumerate}
    \item \textbf{\textsc{Str-EM}} (String Exact Match): the fraction of disambiguated answers that the model includes verbatim in its output.
    \item \textbf{\textsc{Disambig-F1}}: the fraction of answers that can be deduced with a text comprehension model, using the predicted long answer and the disambiguated question:
\end{enumerate}
$$
    \textsc{Disambig-F1} = \frac{1}{N} \sum_k \frac{1}{n^{(k)}}\sum_{i} \phi (y_i^{(k)}, \hat{y}_i^{(k)})
$$
where $N$ is the number of evaluation samples, $n^{(k)}$ the number of disambiguations for the $k$-th question, $y_i^{(k)}$ its $i$-th ground-truth short answer, $\hat{y}_i^{(k)}$ the predicted short answer\footnote{Generated from a RoBERTa~\cite{roberta} model trained on SQUAD2.0~\cite{rajpurkar-etal-2018-know}, with the predicted long answer and the disambiguated question as input.}, and $\phi$ a function that computes the token-level F1 score between them.
    
Finally, they define another metric, namely \textbf{\textsc{DR}}, which is the geometric mean of \textsc{Disambig-F1} and \textsc{Rouge-L}, as an overall estimate of the performance in both disambiguation and answer overlap.

\paragraph{\textbf{Human Evaluation}}
\label{sec:human-metrics}
To better align the model's performance with the overall satisfaction of the human users that interact with the system, we follow \citet{asqa} in creating an anthropocentric evaluation pipeline. Our approach differs in that it performs \textit{head-to-head comparisons} between different models, aiming to draw conclusions about the design choices that lead to a well-performing LFQA system. For a pair of answers, we compare their
\textit{comprehensiveness} (\textsc{Comp}); whether the answer suffices to understand both the source of ambiguity in the question, and the relation between the individual answers, their \textit{fluency} (\textsc{Flue}); whether the answer is coherent and fluent from a human reading stance, and the \textit{overall human impression} (\textsc{Over}); which of the answers is prefered overall.

\section{Results}
\label{sec:results}

\begin{table*}[ht]
    \centering
    \resizebox{\textwidth}{!}{%
        \begin{tabular}{lcccc|c}
            \hline
            & \textbf{\textsc{Answer Length}} & \textbf{\textsc{Rouge-L}} & \textbf{\textsc{Str-EM}} & \textbf{\textsc{Disambig-F1}} & \textbf{\textsc{DR}} \\ \hline
            \textbf{\textsc{Question}} & 71.6 & 15.3 & 1.2 & 0.1 & 1.4 \\ \hline
            \textbf{T5-base \textsc{Closed Book}} & 38.1 & 30.7 & 3.7 & 2.7 & 9.1 \\
            \textbf{BART-base \textsc{Closed Book}} & 44.5 & 31.5 & 3.9 & 2.8 & 9.3 \\
            \textbf{BART-large \textsc{Closed Book}} & 50.2 & 33.4 & 7.1 & 4.5 & \underline{12.2} \\ \hline
            \textbf{BM25@1,3,5} & 103.7 / 310.9 / 518.0 & 28.6 / 20.3 / 15.2 & 18.6 / 30.0 / 36.4 & 10.6 / 14.7 / 17.4 & \textbf{17.4} / 17.3 / 16.2 \\
            \textbf{DPR@1,3,5} & 103.5 / 310.4 / 517.3 & 31.4 / 22.1 / 16.4 & 29.3 / 44.0 / 50.8 & 17.5 / 22.8 / 25.7 & \underline{\textbf{23.5}} / 22.5 / 20.5 \\ \hline
            \textbf{T5-base DPR@1,3,5} & 51.4 / 57.8 / 57.8 & 31.6 / 33.7 / 33.9 & 23.9 / 26.4 / 26.2 & 17.6 / 17.8 / 17.9 & 23.6 / 24.5 / \textbf{24.6} \\
            \textbf{T5-base \texttt{NLGEN} DPR@1,3,5} & 48.5 / 56.0 / 56.0 & 31.2 / 33.7 / 33.7 & 24.6 / 26.0 / 26.7 & 16.8 / 18.1 / 17.8 & 22.9 / \underline{\textbf{24.7}} / 24.5 \\ \hline
            \textbf{BART-base DPR@1,3,5} & 52.4 / 57.1 / 56.7 & \multicolumn{1}{l}{33.1 / 33.9 / 33.9} & \multicolumn{1}{l}{24.2 / 25.1 / 24.5} & \multicolumn{1}{l|}{16.1 / 16.5 / 16.4} & \multicolumn{1}{l}{23.1 / \textbf{23.7} / 23.5} \\
            \textbf{BART-large DPR@1,3,5} & 54.7 / 62.5 / 63.3 & \multicolumn{1}{l}{34.2 / 36.4 / 36.6} & \multicolumn{1}{l}{26.0 / 30.0 / 29.8} & \multicolumn{1}{l|}{18.1 / 20.8 / 20.5} & \multicolumn{1}{l}{24.9 / \textbf{27.5} / 27.4} \\
            \textbf{BART-large \texttt{ELI5} DPR@1,3,5} & 55.2 / 59.7 / 59.9 & 35.1 / 36.6 / 37.0 & 26.7 / 30.3 / 29.7 & 19.0 / 21.0 / 21.0 & 25.8 / \underline{\textbf{27.7}} / 27.6 \\ \hline
        \end{tabular}
    }
    \caption{Performance comparison on the development set of \texttt{ASQA}, using the metrics described in Section~\ref{sec:automated-metrics}. For retrieval-augmented models, $@1,3,5$ indicates using $1/3/5$ retrieved passages as evidence. \textbf{Bold} scores indicate best result among different number of retrieved passages, \underline{underlined} scores indicate best result in each setup.}
    \label{tab:automated-evaluation}
\end{table*}

\paragraph{Automated Evaluation}
We evaluate our models on the development set of \texttt{ASQA} using the automated metrics introduced in Section~\ref{sec:automated-metrics}. Table~\ref{tab:automated-evaluation} displays the results for the naive baseline, the retrieval-only experiments using both BM25 and DPR, and the closed/open book experiments with DPR.
In cases where we use a retriever, we fetch the top-$k$ relevant documents, for $k \in \{1, 3, 5\}$.
As the level of information contained in the retrieved passages has a direct impact on the answer quality, we perform a short study on the upper bound of retrieval in Appendix~\ref{sec:appendix-retrieval-bound}.

Naturally, the naive \textsc{Question} baseline performs the worst, with a \textsc{Rouge-L} score of 15.3 and a \textsc{Disambig-F1} score of 0.1, leading to an overall DR score of 1.4. Focusing on the retrieval-only experiments, we observe that DPR consistently outperforms BM25 in all metrics, which is anticipated as semantic matching allows us to retrieve passages that answer different aspects of the question. Remarkably, although the closed-book variants surpass the retrieval-only methods in terms of \textsc{Rouge-L}, we see the opposite trend for all other metrics. The open-book variants outperform the rest, confirming our hypothesis that augmenting the generative model with a retriever is crucial for performance. We also notice that the increase in performance closely follows the growth in parameter count, implying that \textbf{[RQ1]} model scaling improves the quality of the answer.\footnote{To gain further insight into some of the common errors, we showcase a few example model outputs in Appendix~\ref{sec:appendix-answers}.}
Surprisingly, the impact of fine-tuning on larger collections on performance is marginal, as the disambiguation metrics only narrowly improve. For completeness, we report the intermediate fine-tuning results for the closed-book experiments in Appendix~\ref{sec:appendix-pre-training-closed-book}. In general, we deem that \textbf{[RQ2]} naively fine-tuning on non-ambiguous LFQA datasets has limited added value, as the models are not inclined to identify any uncertainty in the meaning of the questions asked.

\paragraph{Human Evaluation}
We perform a head-to-head comparison between a selection of approaches to determine which modeling aspects contribute more/less to the performance. For each pair, we asked 2 assessors to compare the models \textit{blindly} using the human evaluation metrics defined in Section~\ref{sec:human-metrics}. We report our results in Table~\ref{tab:head-to-head}, where we observe that most comparisons follow the trends we described previously using the automated metrics. Noticeably, when comparing T5-base DPR@3 vs BART-base DPR@3 and DPR@1 vs BART-base DPR@1, human annotators seem to equally prefer both models overall. However, the difference in \textsc{DR} for these pairs is merely a few decimal points, which justifies the inability to distinguish between them. Hence, we confirm that \textbf{[RQ3]} the \textit{overall human impression} aligns with automated metrics.

\begin{table}[t]
    \centering
    \resizebox{\columnwidth}{!}{%
            \begin{tabular}{lccc}
            \hline
            & \textbf{\textsc{Comp}} & \textbf{\textsc{Flue}} & \textbf{\textsc{Over}} \\ \hline
            \rule{0pt}{\normalbaselineskip} \\[-\normalbaselineskip]
            \textbf{T5-base} vs \textbf{BART-base}$^*$ &
            \comp{0.5} & \flue{0.38} & \overall{0.5} \\
            \textbf{BART-large} vs \textbf{BART-base}$^*$ & 
            \comp{0.62} & \flue{0.62} & \overall{0.62} \\
            \textbf{BART-large \texttt{ELI5}} vs \textbf{BART-large}$^*$ & 
            \comp{0.62} & \flue{0.5} & \overall{0.62} \\
            \textbf{BART-large: DPR@3} vs \textbf{DPR@1} & 
            \comp{0.75} & \flue{0.5} & \overall{0.75} \\
            \textbf{DPR@1} vs \textbf{BART-base DPR@1} & 
            \comp{0.62} & \flue{0.38} & \overall{0.5} \\
            \textbf{BART-large DPR@1} vs \textbf{DPR@1} & 
            \comp{0.38} & \flue{0.5} & \overall{0.75} \\
            \hline
            \end{tabular}
        }
        \caption{Head-to-head comparison using human evaluation metrics. Pairs marked with $^*$ use DPR@3 for retrieval. Bars indicate the annotators' preference for either model.}
        \label{tab:head-to-head}
        \vspace{-.45cm}
\end{table}

\paragraph{Random Retrieval}
Finally, we evaluate whether the generated answers are grounded on the retrieved passages. Table~\ref{tab:random-retrieval} showcases the performance for the models used in our open-book experiments, and we generally see a \textit{decrease} in performance when compared to fetching relevant documents. Interestingly, BART-base with random retrieval performs better compared to its closed-book variant. As this is the smallest model we tested, we speculate that for LFQA, under-parameterized architectures may be relying more on the question and not the corresponding evidence. Despite this, \textbf{[RQ4]} models for ambiguous LFQA appear to be well grounded in the retrieved evidence overall.

\begin{table}[t]
    \centering
    \resizebox{\columnwidth}{!}{%
        \begin{tabular}{lccc|c}
            \hline
             & \textbf{\textsc{Rouge-L}} & \textbf{\textsc{Str-EM}} & \textbf{\textsc{Dis-F1}} & \textbf{\textsc{DR}} \\ \hline
            \textbf{T5-base} & 20.5 & 2.0 & 0.7 & 3.7 \\
            \textbf{T5-base \texttt{NLGEN}} & 20.8 & 1.6 & 0.8 & 4.1 \\
            \textbf{BART-base} & 29.8 & 5.5 & 3.5 & 10.3 \\
            \textbf{BART-large} & 24.3 & 4.4 & 2.2 & 7.2 \\
            \textbf{BART-large \texttt{ELI5}} & 29.6 & 4.8 & 3.0 & 9.5\\
            \hline
        \end{tabular}
    }
    \caption{Random Retrieval performance. We use $k=3$ random passages from the DPR index as evidence.}
    \label{tab:random-retrieval}
\end{table}
\section{Conclusion}
\label{sec:conclusion}
In conclusion, by exploring the LFQA field in answering ambiguous questions, we find that the \texttt{ASQA} dataset is a good foundation to develop and evaluate models. We notice that larger generative models produce better answers, with semantic matching for retrieval having a positive impact. To compensate for the small size of the dataset, we experiment with intermediate fine-tuning on larger collections and find that doing so only marginally improves the results. By comparing the performance when using relevant versus random documents, we show the models' dependency on the provided context. Our human evaluation confirms the general trends we observed using the automated metrics, giving credit to their disambiguating ability.

\section*{Limitations}
We see two main limitations in our study. Primarily, given the clear trend of larger generative models producing higher quality answers, an obvious question is to investigate whether this continues to be the case indefinitely, or whether it saturates after a critical amount of parameters. Despite this, due to hardware restrictions, we were unable to experiment with models larger than BART-large. Additionally, considering that the field of ambiguous QA inherently requires complementary pieces of evidence, there is no doubt that diversification methods are bound to yield better results in terms of disambiguation quality. In this work, however, we limited ourselves to using a typical neural retriever, shifting our focus toward the factuality and the fluency of the generated answers.

\section*{Ethics Statement}
As we use a publicly available Wikipedia index as our knowledge base, it is possible that the generated answers may contain some form of bias that reflects the information submitted by the anonymous editors of the website. To prevent this, follow-up work could examine how to detect misinformation or hate speech in indexed passages, before using them as evidence for the generative models.

\section*{Acknowledgements}
This work was carried out on the Dutch national e-infrastructure with the support of the SURF Cooperative. We would also like to thank Evangelos Kanoulas for his valuable feedback and guidance during this project.

\bibliography{anthology,main}
\bibliographystyle{acl_natbib}

\clearpage
\appendix
\section{Potential Issues with \texttt{ASQA}}
\label{sec:appendix-asqa-issues}
During our experiments, we notice certain issues regarding the \texttt{ASQA} dataset. In particular, we observed that in some cases the ambiguity in the question is pedantic. For instance, there are questions where a date is not specified (e.g., ``What kind of car won the Daytona 500 this year?''), and in a typical scenario, the system would assume that the question refers to the current year. Instead, in \texttt{ASQA} this type of question is considered ambiguous, and the short answers resolve the ambiguity by reformulating the question for different years (e.g., 2017, 2016, 2015, etc.). In addition, we argue that some of the disambiguated questions and their short answers are too specific. This results in the model being penalized for not generating the exact correct terms, even though its answer semantically covers some part of that interpretation (e.g., the last example in Table~\ref{tab:answer-analysis}). 

\section{Implementation Details}
\label{sec:appendix-training}
We use the official T5-base, BART-base, and BART-large implementations from HuggingFace. We train each model for $20$ epochs on the \texttt{ASQA}\footnote{https://huggingface.co/datasets/din0s/asqa} dataset with the \texttt{AdamW} optimizer~\cite{adamw}, using a weight decay of $0.01$, and a learning rate of $10^{-5}$ for T5 and $5\cdot10^{-6}$ for BART. Training is stopped early if the validation loss doesn't decrease after $5$ epochs. We use a train batch size of $8$ for open-book experiments, and $16$ for closed-book, with an evaluation batch size of $8$ in both cases. We train our models on one NVIDIA Titan RTX GPU, and use 16-bit mixed precision to accelerate training. All of the models converged within $\sim$30 minutes of training.

For our intermediate fine-tuning experiments, we first train T5-base on \texttt{NLGEN} subset\footnote{https://huggingface.co/datasets/din0s/msmarco-nlgen} of the \texttt{MS MARCO} dataset for $1$ epoch with a learning rate of $10^{-4}$ and then continue training on \texttt{ASQA} as described above. For BART, we use a publicly available instance from HuggingFace that has been pre-trained on \texttt{ELI5}\footnote{https://huggingface.co/vblagoje/bart\_lfqa}, and continue training on \texttt{ASQA}.

Finally, we use beam decoding with $5$ beams and a max sequence length of $100$ tokens. We force the model to not repeat the same trigram in its output by using the option \texttt{no\_repeat\_ngram\_size=3} when generating answers.

\section{Upper Bound for Retrieval}
\label{sec:appendix-retrieval-bound}
To investigate the efficiency of our retrieval models, we perform an analysis for the upper bound of the retrieved passages' relevancy to the corresponding gold answers. In particular, for each of the systems used, we count the number of relevant short answers retrieved using an exact string match with the dataset's gold answers. In Table~\ref{tab:retrieval-upper-bound}, we notice that even the best retrieval setup at our disposal, namely \textbf{DPR@5}, only fetches $\sim44\%$ of the relevant answers on average or up to $58\%$ in cases where it identifies at least one relevant passage. This emphasizes the need for a multi-hop retrieval system like JPR~\cite{jpr} in order to fully utilize the power of the first component of the pipeline. It is evident that a higher upper bound for retrieval will increase the overall performance of the systems that tackle ambiguous LFQA, making it a high priority for follow-up work.

\begin{table}[ht]
    \centering
    \begin{tabular}{@{}lcc@{}}
        \hline
         & \multicolumn{2}{c}{\textbf{Avg \# of short answers retrieved}} \\ \hline
         & In all results & \begin{tabular}[c]{@{}c@{}}In results with\\ $\geq1$ correct answer\end{tabular} \\ \hline
        \textbf{BM25@1} & 0.52 (14.71\%) & 1.61 (45.44\%) \\
        \textbf{BM25@3} & 0.89 (24.10\%) & 1.83 (49.35\%) \\
        \textbf{BM25@5} & 1.11 (29.81\%) & 1.96 (52.53\%) \\ \hline
        \textbf{DPR@1} & 0.89 (24.70\%) & 1.74 (48.38\%) \\
        \textbf{DPR@3} & 1.39 (37.73\%) & 2.00 (54.61\%) \\
        \textbf{DPR@5} & 1.62 (43.79\%) & 2.15 (58.15\%) \\ \hline
    \end{tabular}
    \caption{Number of relevant chunks of evidence identified using different retrieval systems. This constitutes an upper bound to the generative model's ability to answer the various disambiguated questions.}
    \label{tab:retrieval-upper-bound}
    \vspace{-.55cm}
\end{table}

\section{Effect of Intermediate Fine-Tuning for the Closed Book Experiments}
\label{sec:appendix-pre-training-closed-book}
Table~\ref{tab:pre-training-closed-book} displays the impact of the intermediate fine-tuning on a larger LFQA collection in the closed-book setting, using automated evaluation metrics. Similarly to the open-book setting, \textbf{BART-large \texttt{ELI5}} performs the best, which confirms the hypothesis that larger models are able to benefit the most when given more training data.

\begin{table}[h]
    \centering
    \resizebox{\columnwidth}{!}{%
        \begin{tabular}{lccc|c}
            \hline
             & \textbf{\textsc{Rouge-L}} & \textbf{\textsc{Str-EM}} & \textbf{\textsc{Dis-F1}} & \textbf{\textsc{DR}} \\ \hline
            \textbf{T5-base} & 30.7 & 3.7 & 2.7 & 9.1 \\
            \textbf{T5-base \texttt{NLGEN}} & 30.0 & 2.9 & 2.8 & 9.2 \\
            \textbf{BART-base} & 31.5 & 3.9 & 2.8 & 9.3 \\
            \textbf{BART-large} & 33.4 & 7.1 & 4.5 & 12.2 \\
            \textbf{BART-large \texttt{ELI5}} & 34.1 & 7.7 & 4.8 & 12.8
            \\ \hline
        \end{tabular}
    }
    \caption{Closed-book performance, showcasing the effect of intermediate fine-tuning on a large LFQA dataset.}
    \label{tab:pre-training-closed-book}
    \vspace{-.55cm}
\end{table}

\section{Answer Demonstration}
\label{sec:appendix-answers}
We demonstrate a brief selection of answers generated by our best-performing model (\textbf{BART-large \texttt{ELI5}}) in Table~\ref{tab:answer-analysis}. We observe that the main reason for our model's inability to disambiguate all parts of the question is that the passages retrieved with DPR simply do not cover all possible interpretations. In most cases, the first passage retrieved suffices to answer at least one of the disambiguated questions, and the rest passages are either not relevant or answer the same question again. Contrary to the results of \citet{asqa}, our model does not hallucinate or repeat itself. We hypothesize that since their model (\textbf{T5-large}) is almost twice as big as \textbf{BART-large} in terms of parameter count, it can better utilize its context but at the same time be overconfident in its answers.

\begin{table*}[ht]
\centering
\footnotesize
\vspace{-0.5cm}
\begin{tabular}{p{3.5cm}|p{3.5cm}|p{4.5cm}|p{2.75cm}}
\toprule
 \textbf{Question, provided disambiguations, and answers} & \textbf{Generated model output: BART-large \texttt{ELI5} DPR@3} & \textbf{DPR@3 retrieved passages} & \textbf{Answer analysis} \\
\midrule
\textit{Q}: Who was the ruler of France in 1830?
\newline
\newline
\textit{Q1}: Who was the ruler of France until 2 August 1830? \newline
\textit{A1a}: Charles Philippe
\newline
\textit{A1b}: Charles X
\newline
\newline
\textit{Q2}: Who was the ruler of France after 9 August 1830? \newline
\textit{A2}: Louis-Philippe I
\newline
&
the french revolution of 1830, also known as the july revolution or in french, the second french revolution, led to the overthrow of \textcolor{mygreen}{king charles x}, the french bourbon monarch, and the ascent of his cousin \textcolor{mygreen}{louis philippe}, duke of orléans, who himself, after 18 precarious years on the throne, would be overthrown in 1848.
& 
\textbf{\textit{July Revolution}} The French Revolution of 1830, also known as the July Revolution ("révolution de Juillet"), Second French Revolution or in French ("Three Glorious [Days]"), led to the overthrow of King Charles X, the French Bourbon monarch, and the ascent of his cousin Louis Philippe, Duke of Orléans, who himself, after 18 precarious years on the throne, would be overthrown in 1848.. 

\textbf{\textit{July Monarchy}} 1840, to general indifference. The public\'s attention was concentrated on the trial of Marie Lafarge..

\textbf{\textit{July Revolution}}
..led to the overthrow of King Charles X, the French Bourbon monarch, and the ascent of his cousin Louis Philippe, Duke of Orléans, who himself..

&
We do not notice any hallucination or repetition in this example. The model manages to correctly answer the ambiguous question with fluency and coherency. In addition, through reading the answer we understand why the question asked was ambiguous. \\ 
\midrule

\textit{Q}: The repetition of phrases or sentences with similar structures or meanings is called? \newline
 \newline
\textit{Q1}:The repetition of phrases or sentences with similar structures is called? \newline
\textit{A1a}: parallel construction
\newline
\textit{A1b}: parallelism
\newline
\textit{A1c}: Parallel syntax
\newline
\newline
\textit{Q2}: The repetition of phrases or sentences with similar meanings is called? \newline
\textit{A2}: Schesis onomaton
\newline &
\textcolor{mygreen}{parallelism, also known as parallel structure or parallel construction, is a balance within one or more sentences of similar phrases or clauses that have the same grammatical structure}. the application of parallelism affects readability and may make texts easier to process.
&
\textbf{\textit{Parallelism (grammar)}} In grammar, parallelism, also known as parallel structure or parallel construction, is a balance within one or more sentences of similar phrases or clauses that have the same grammatical structure..

\textbf{\textit{Metathesis (linguistics)}} Metathesis (; from Greek , from "I put in a different order"; Latin: "trānspositiō") is the transposition..

\textbf{\textit{Resultative}} Resultative In linguistics, a resultative (abbreviated ) is a form that expresses that something or someone has undergone a change in state as the result of the completion of an event..
& 
In this example, the model only manages to get one of the two answers that disambiguate the question. We do not observe any hallucinations or repetition.\newline

 \\ 
 \midrule
 \textit{Q}: When does callie come in grey's anatomy? \newline
\newline 
\textit{Q1}: What date does Callie Torres first appear in Grey's Anatomy? \newline
\textit{A1}: February 26, 2006 \newline
\newline
\textit{Q2}: Which episode does Callie Torres first appear in Grey's Anatomy? \newline
\textit{A2}: "What Have I Done to Deserve This?" (2.19)
\newline
\newline
\textit{Q3}: What date does Callie Torres start to appear in Grey's Anatomy as a series regular?\newline
\textit{A3}: September 21, 2006 \newline
\newline
\textit{Q4}: What episode Callie Torres start to appear in Grey's Anatomy as a series regular? \newline
\textit{A4}: "Time Has Come Today" (3.01) \newline
&
grey's anatomy is an american medical drama television series created by shonda rhimes, which airs on the abc in the united states. the character \textcolor{myorange}{callie torres was introduced in season two}, as a senior orthopedic resident as a love interest for intern george o'malley (t.r. knight). \textcolor{myorange}{she was originally contracted to appear on a recurring basis, but received star billing in the third season.} at the end of the show's twelfth season, ramirez
&
\textbf{\textit{Callie Torres}} ..."Callie" Torres, M.D. is a fictional character from the medical drama television series "Grey\'s Anatomy"... She was introduced in season two, as a senior orthopedic resident, as a love interest for intern George O\'Malley (T.R. Knight). Eventually becoming an attending orthopedic surgeon, the character was originally contracted to appear on a recurring basis, but received star billing..

\textbf{\textit{Callie Torres}} ..Eventually becoming an attending orthopedic surgeon, the character was originally contracted to appear on a recurring basis, but received star billing in the third season..'

\textbf{\textit{Callie Torres}} Ramirez was nominated for several awards for her portrayal of Torres, including the..

&
Here the model is not able to get any of the answers correctly. However, one could argue that two of the four disambiguated questions were partially answered since the model outputs that the character was introduced in season two and became a series regular in the third season. We also do not notice any hallucinations or repetition.
 \\ 

\bottomrule
\end{tabular}
\caption{\label{tab:examples} Qualitative analysis. Green/orange text highlights correct/partially-correct parts of the answer.}
\label{tab:answer-analysis}
\end{table*}

\end{document}